\documentclass[letterpaper, 10 pt, conference]{ieeeconf}  

\IEEEoverridecommandlockouts                              

\usepackage{graphicx}
\usepackage{xcolor}
\usepackage{fancyhdr}

\makeatletter
\def\endfigure{\end@float}
\def\endtable{\end@float}
\makeatother

\usepackage[caption=false]{subfig}
\usepackage{ctable} 

\title{\LARGE \bf
Two Eyes Are Better Than One: Exploiting  Binocular Correlation for Diabetic Retinopathy Severity Grading}

\author{Peisheng Qian$^{*1}$, Ziyuan Zhao$^{*1}$, Cong Chen$^{1,2}$, Zeng Zeng$^{\dagger 1}$, Xiaoli Li$^{1}$
\thanks{* Contributed equally. $^{\dagger}$Corresponding author. This research is supported by Institute for Infocomm Research (I2R), Agency for Science, Technology and Research (A*STAR), Singapore. $^{1}$ Institute for Infocomm Research (I2R), Agency for Science, Technology and Research (A*STAR), Singapore. $^{2}$ National University of Singapore, Singapore. This work was done when Cong Chen was an intern at I2R, A*STAR.}
}

\begin{document}
\renewcommand{\thefigure}{\arabic{figure}}
\maketitle
\thispagestyle{empty}
\pagestyle{empty}


\thispagestyle{fancy}
\fancyfoot{}
\lfoot{\scriptsize{© 2021 IEEE.  Personal use of this material is permitted.  Permission from IEEE must be obtained for all other uses, in any current or future media, including reprinting/republishing this material for advertising or promotional purposes, creating new collective works, for resale or redistribution to servers or lists, or reuse of any copyrighted component of this work in other works.}}

\begin{abstract}
\textit{ Diabetic retinopathy (DR)} is one of the most common eye conditions among diabetic patients. However, vision loss occurs primarily in the late stages of DR, and the symptoms of visual impairment, ranging from mild to severe, can vary greatly, adding to the burden of diagnosis and treatment in clinical practice. Deep learning methods based on retinal images have achieved remarkable success in automatic DR grading, but most of them neglect that the presence of diabetes usually affects both eyes, and ophthalmologists usually compare both eyes concurrently for DR diagnosis, leaving correlations between left and right eyes unexploited. In this study, simulating the diagnostic process, we propose a two-stream binocular network to capture the subtle correlations between left and right eyes, in which, paired images of eyes are fed into two identical subnetworks separately during training. We design a contrastive grading loss to learn binocular correlation for five-class DR detection, which maximizes inter-class dissimilarity while minimizing the intra-class difference. Experimental results on the EyePACS dataset show the superiority of the proposed binocular model, outperforming monocular methods by a large margin.

\indent \textit{Clinical relevance}— Compared to conventional DR grading methods based on monocular images, our approach can provide more accurate predictions and extract graphical patterns from retinal images of both eyes for clinical reference. 
\end{abstract}

\section{INTRODUCTION}\label{section1}

Diabetic retinopathy (DR) is one of the most prevailing eye diseases among patients with diabetes. It has become the primary cause of blindness in the working-age population of the developed world~\cite{Das_2018}. In Singapore, over $40\%$ of diabetic patients suffer from DR at various stages from mild to severe~\cite{2020SEA}. Prevention of DR is challenging because the symptoms of DR are hardly recognizable at the early stage. The gold standard for diagnosis of DR is digital color fundus photography. Digital color fundus photography is the gold standard for diagnosing DR. However, observing and evaluating fundus images is time-consuming and labor-intensive, requiring experienced ophthalmologists.


\begin{figure}[t]
    \centering
    \includegraphics[width = 8.3cm]{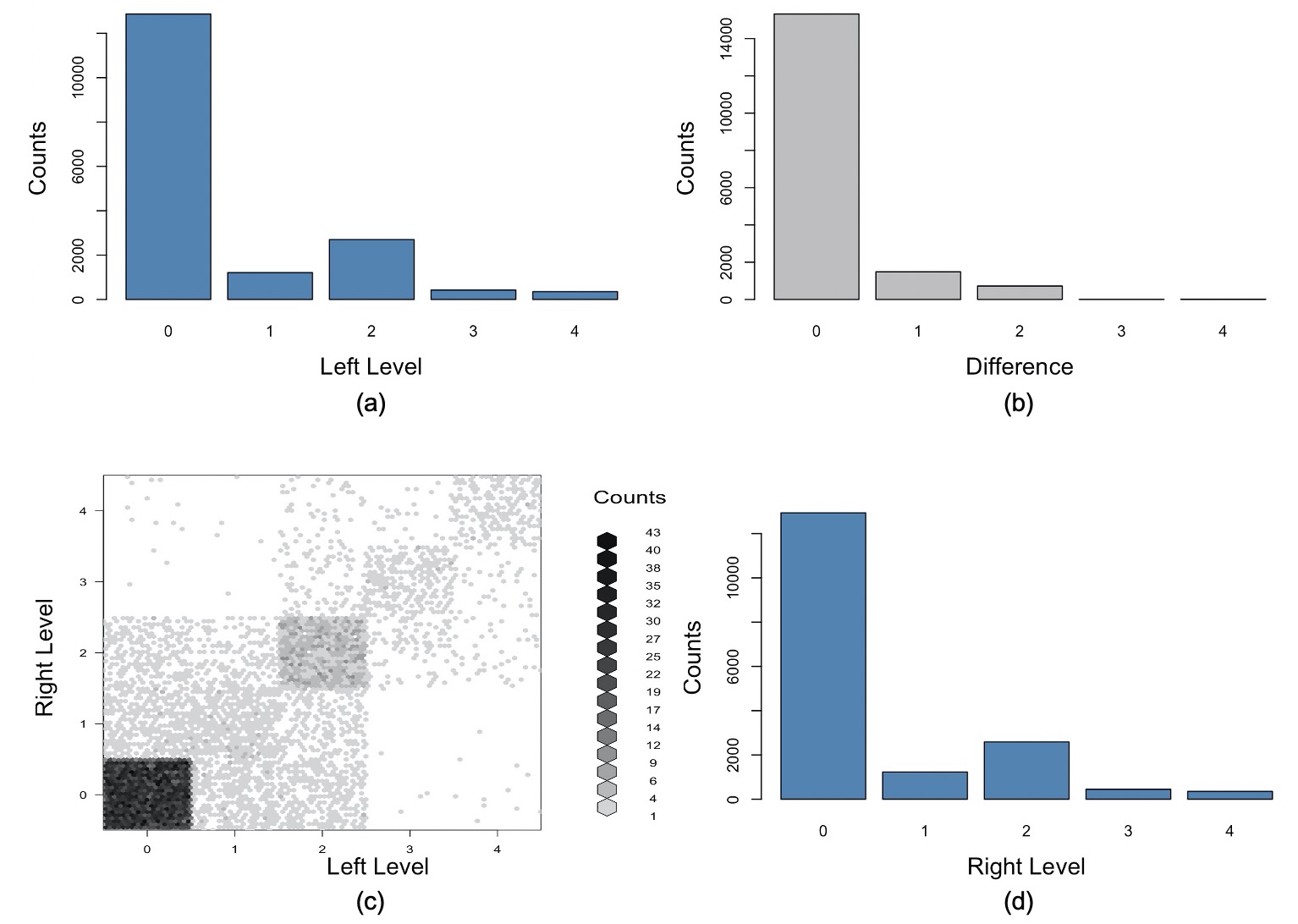}
    \caption{(a) The distribution of $C_l$ (disease severity levels of left eyes). (b) The distribution of $|C_l-C_r|$ (difference between disease severity levels of each patient's left and right eyes). (c) The scatter plot of disease severity levels between each patient's left and right eyes. (d) The distribution of $C_r$ (disease severity levels of right eyes).}
    \label{fig:EDA}
\end{figure}



Deep learning approaches have achieved great success in DR grading based on retinal images~\cite{li2021reiew}. Different from conventional machine learning methods, which rely on hand-crafted features,~\emph{e.g.}, the retinal blood vessels and the optic disc, deep neural networks can effectively extract high-level features and learn complex representations from retinal images, which better facilitates the clinical process and eliminates human errors. Most of the existing methods take monocular images as the model inputs, regardless of the difference between left and right eyes~\cite{2020SEA, wang2017zoom, zhao2019biranet}. However, in the clinical diagnosis of eye diseases, both the left and right eyes are taken into consideration~\cite{liu2005variation, Eppig2018}. In other words, the correlation between left and right eyes can be used for DR grading in clinical practice. As shown in Fig.~\ref{fig:EDA}, we performed exploratory data analysis on the Kaggle dataset provided by EyePACS~\cite{kaggle}. For better visualization, we add random variance to each level of two eyes concurrently in Fig.~\ref{fig:EDA}~(c). We can see that both eyes of the same patient are highly correlated and the Pearson’s correlation coefficient $\rho$ of them is 0.85~\cite{kaggle}. Motivated by the clinical process and our analysis, we hypothesize that the left-right correlation can be used in deep learning for better DR grading.


In this work, we propose a two-stream binocular network. The network consists of $2$ convolutional neural networks (CNNs) that share the same weights and take the left and right eyes of the same patient as inputs, respectively. The model is simultaneously updated by both left and right eye images. With this learning framework, the model can recognize individual patterns in each eye as well as the similarities between the left and right eyes for DR grading. We propose a hybrid loss to optimize the network. The loss consists of a contrastive grading loss and a weighted cross-entropy loss. More specifically, we introduce the contrastive grading loss to optimize the network in accordance with the left-right similarity, in which we scale the contrastive loss proportional to the discrepancies in DR grading for finer classification granularity. The contribution of this paper are summarized as follows: 

\begin{itemize}
    \item We construct a two-stream binocular network that consists of two identical networks with shared weights. We design a novel training strategy that takes both left and right eyes of the same patient as inputs. 

    \item We propose a hybrid loss function which consists of the contrastive grading loss and the weighted cross-entropy loss. The contrastive grading loss optimizes the network based on the similarity between the left and right eyes. Experiments show that with the proposed loss function, the network can recognize similarities among left and right eyes, and obtain superior results than baselines.
\end{itemize}


\section{RELATED WORK}\label{section2}
Early studies relied heavily on experts manually extracting features and certain textural properties for DR classification~\cite{ahmad2014image}. In recent years, deep learning techniques, such as CNNs, have been proved effective in DR grading. Bravo~\emph{et al.} designed a VGG-based network architecture and combine it with various pre-processing images~\cite{Mar2017Automatic}, reaching $50.5\%$ classification accuracy on a balanced dataset. Zhao~\emph{et al.} described a bilinear model with an attention mechanism for fine-grained classification of DR~\cite{zhao2019biranet}. Wang~\emph{et al.} implemented Zoom-in-Net with multiple sub-networks for DR grading and localization of suspicious regions~\cite{wang2017zoom}. Zhao~\emph{et al.} further investigated the subtle differences between different DR severity levels and developed a new network architecture, SEA-Net, in which spatial and channel attention are alternatively stacked~\cite{2020SEA}. The aforementioned methods enhance model architectures and prove their effectiveness in DR grading. However, they do not leverage the left-right eye correlations. 

Existing research suggests that the correlation between left and right eyes could be explored on eye symptoms~\cite{liu2005variation, Eppig2018, zeng2019automated}. We are not the first one to address the correlation between the two eyes. Zeng~\emph{et al.} presented promising results with binocular inputs to siamese-like deep learning models for DR classification~\cite{2020SEA}. While this method introduces weights sharing between CNNs in their model, it omits the calculation of similarities or variances among different DR grades. There is also no modification to the original contrastive loss. To overcome the above-mentioned shortcomings and clearly reflect the differences between DR grades in the loss function, we propose a two-stream binocular network and a novel contrastive grading loss, which are illustrated in Section~\ref{section3}.

\section{METHODOLOGY}
\label{section3}

\begin{figure}[t]
    \centering
    \includegraphics[width = 8.3cm]{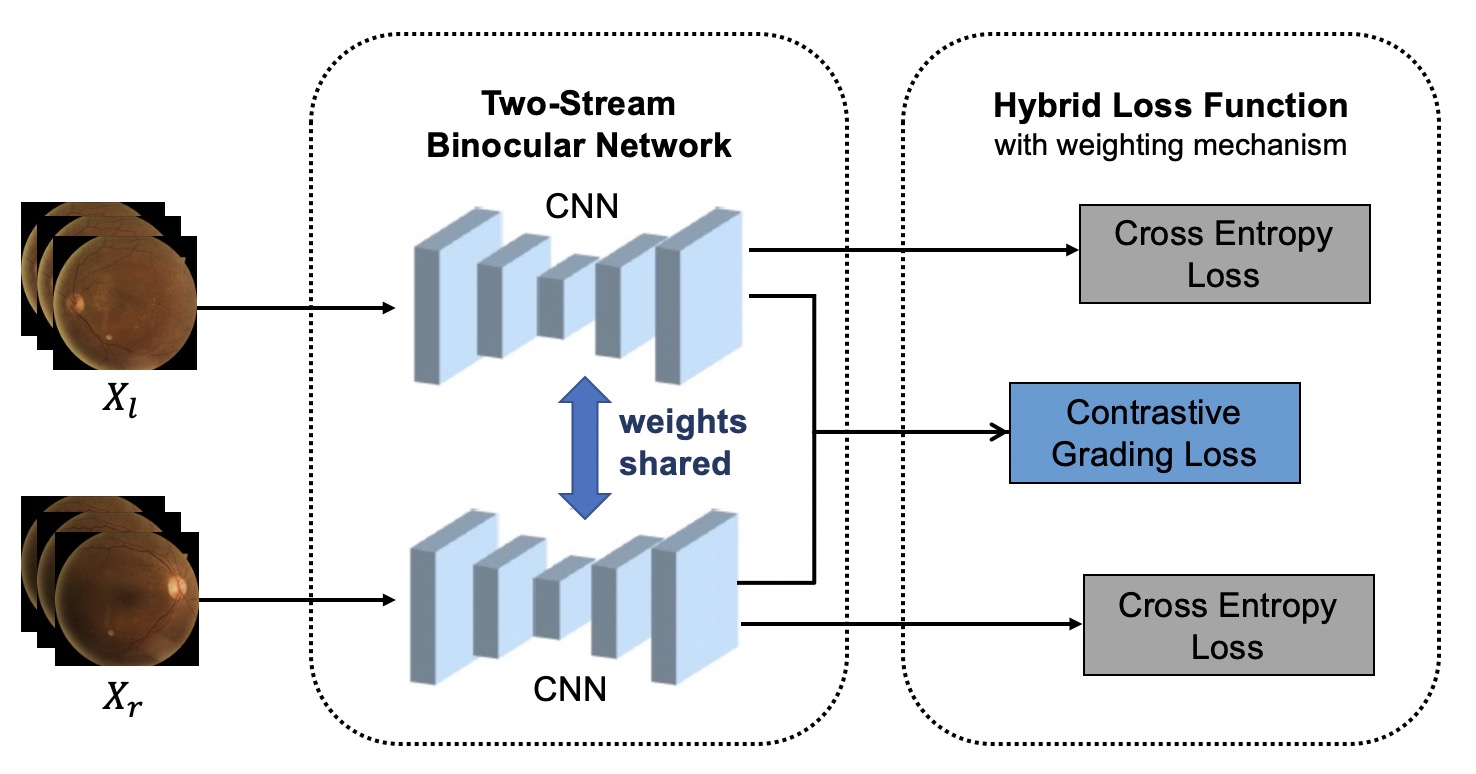}
    \caption{The architecture of the two-stream binocular network.}
    \label{fig:Architecture}
\end{figure}

The proposed two-stream binocular network and training strategy for DR grading is illustrated in Fig.~\ref{fig:Architecture}, in which, a pair of left and right eye images {$X_l$, $X_r$} are taken as inputs to two identical sub-networks respectively. The two sub-networks with shared weights extract features from the two eyes and classify their DR grading separately. To leverage the left-right eye correlations, we apply a novel loss function during training, which includes a contrastive grading loss and a weighted cross-entropy loss.

\subsection{Two-stream Binocular Network}
The proposed framework consists of two convolutional neural networks (CNNs) with shared weights. In this study, we use ResNet-50~\cite{ResNet} and BiRA-Net~\cite{zhao2019biranet} as the backbones of the network to study the effectiveness of the proposed architecture. Both eyes from the same patient are paired and passed to each of the CNNs. The network captures features from both left and right eyes as well as the similarities between them, based on which the network classifies DR grading for both eyes. The process simulates the real-life clinical DR diagnosis on both eyes and utilizes the correlations between them for diagnosis.



\subsection{The Proposed Hybrid Loss}
\subsubsection{Contrastive grading loss}
To optimize the network and extract similarities between left and right eyes, we propose the contrastive grading loss. The loss function is improved from the contrastive loss, which is commonly used in conjunction with siamese networks~\cite{li2020siamese}. The loss calculates the similarity between eyes based on Euclidean distances between hidden features in the last layers of the $2$ CNNs. Assuming that the disease severity levels of paired images $X_l$ and $X_r$ are $C_l$ and $C_r$ respectively, the grading difference between the paired images can be represented as $|C_l-C_r|$.  The contrastive grading loss function is defined as:
\begin{equation}
    \mathcal{L}_{cg}={F}_{c}d^{2}+ |C_l-C_r|{max(margin-d, 0)}^{2}
    \label{eqn:contrastive_loss}
\end{equation}
where $F_c$ is the binary control factor for the first term in the loss function.  If the disease severity level in both images is the same, $F_c=1$. Otherwise, $F_c=0$. $d$ represents the Euclidean distance between outputs from the two sub-networks in the two-stream binocular network. $margin$ is the threshold, which is adjusted empirically and set to $2$ in this study~\cite{li2020siamese}. Compared to the existing contrastive loss, Eqn.~\ref{eqn:contrastive_loss} scales the second term in the loss function proportional to the differences between left and right features and therefore optimizes the network to recognize similarities between left and right eyes in the feature space.
\subsubsection{Weighted Cross-Entropy Loss}
To alleviate the overfitting problem due to the imbalance of the DR dataset, we add the weighting mechanism in our hybrid loss function. $x[0],  x[1], x[2], \dots, x[C-1]$ denotes the class probability of input $x$, and the class index $y$ is in the range $[0,C-1]$, where $C$ is the number of classes. Each sample is scaled by the weight proportional to the inverse of the percentage of the class of the sample in the training set, denoted as $weight[y]$ in Eqn.~\ref{eqn:wce}. The weighted cross-entropy loss~\cite{article} is formulated as:

\begin{equation}
    \mathcal{L}_{ce}=weight[y]\left(-x[y] +\log\left (\sum_{i=0}^{C-1}\exp(x[i]) \right )\right)
    \label{eqn:wce}
\end{equation}

Finally, the proposed hybrid loss function is a weighted sum of the contrastive grading loss and the cross-entropy loss, which is defined in Eqn.~\ref{eqn:hybrid}:
\begin{equation}
    \mathcal{L}=\lambda\mathcal{L}_{cg}+\mathcal{L}_{ce}
    \label{eqn:hybrid}
\end{equation}
where $\lambda$  is the factor controlling the scale of the contrastive grading loss in the hybrid loss. In the experiments, $\lambda$ is empirically set to $0.1$ for model optimization.

\section{EXPERIMENTS}
\label{section4}
\subsection{Dataset and Implementation}
We collate the dataset provided by EyePACs, hosted on Kaggle~\cite{kaggle}. The dataset is labeled with a set of definitions to ensure label consistency. The grades of DR are categorized into $5$ classes from $0$ to $4$ with increasing disease severity. Grade $0$ indicates non-diabetic, while grade $4$ indicates the most severe diabetes. We randomly split the retinal images from the dataset into $33,566$ images as the training set and $1,560$ images as the test set. The test set is balanced.



The implementation details are described as follows. Left and right retina images of the same patients are selected in pairs as the input. To augment the training set, random horizontal and vertical flipping, and random rotation of $\pm10$ degrees are applied to the input images. The images are then resized to $224\times224$. Finally, the images are standardized across the RGB channels by subtracting the mean and dividing by the standard deviation of each channel. 

We load the ImageNet pre-trained weights into the network before starting the training process~\cite{ResNet}. The network is trained using the stochastic gradient descent (SGD) optimizer with an initial learning rate of $0.001$ and a weight decay factor of $1e-8$. The learning rate is multiplied by $0.1$ when the model performance on the test set does not improve for $10$ consecutive epochs. The network is trained for $100$ epochs with a batch size of $32$. The experiments are implemented on NVIDIA RTX 2080Ti GPUs with Pytorch 1.7.1.

\subsection{Performance Metrics}
For a comprehensive evaluation, we employ $3$ commonly-used metrics to quantitatively evaluate the performance, which have also been used in previous research~\cite{2020SEA, zhao2019biranet}. They are:
\begin{itemize}
    \item ACA: Average classification accuracy.
    \item F1: Averaged Macro-F1 score of the 5 classes.
    \item AUC: The area under the receiver operating characteristics (ROC) curve.
\end{itemize}
\subsection{Baseline Methods}
\label{sec:baseline}
We compare our framework with several baselines. In~\cite{Mar2017Automatic}, a VGG-based classifier is trained on the dataset with preprocessing techniques including circular RGB, grayscale and color-centered sets. Zhao~\emph{et al.} report results of ResNet-50 on the Kaggle dataset, and we re-implement ResNet-50 with slightly higher ACA~\cite{zhao2019biranet}. We combine ResNet-50 with mean squared error (MSE) in the loss function as another baseline. In~\cite{zhao2019biranet}, BiRA-Net is invented, which features a bilinear learning strategy together with a grading loss for fine-grained classification. Our methods with different backbones are shown as follows, 
\begin{itemize}
\item TSBN (ResNet-50): two-stream binocular network, with ResNet-50 as the backbone.
\item TSBN (BiRA-Net): two-stream binocular network, with BiRA-Net as the backbone. The network consists of $2$ BiRA-Net models that share weights.
\end{itemize}

\subsection{Results and Discussion}

\begin{figure*}[ht]
    \centering
    \includegraphics[width=\textwidth]{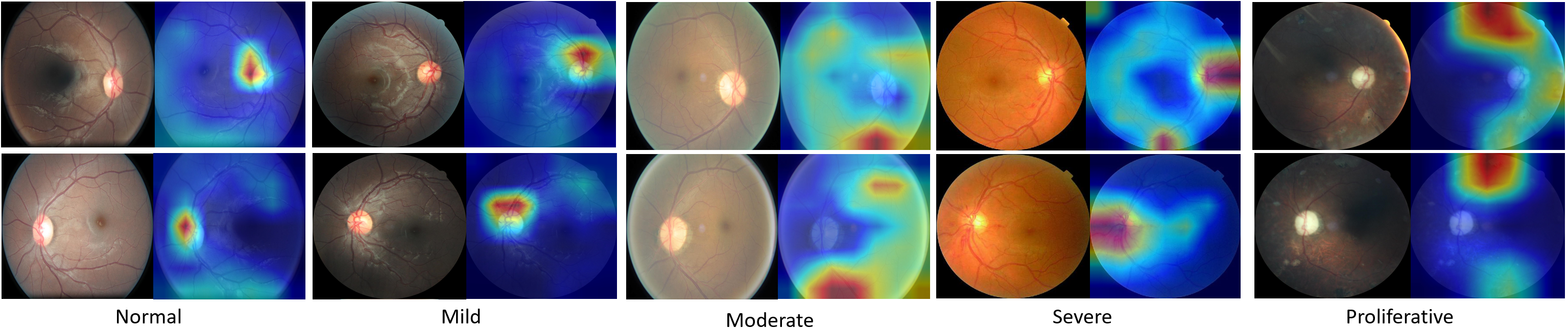}
    \caption{Illustrative examples of original images and saliency maps from TSBN (ResNet-50). Each column contains left (top) and right (bottom) retina images from the same patient who has the same DR level in both eyes.}
    \label{fig:gradcam}
\end{figure*}

\begin{table}[htbp]
\caption{Experimental results on DR grading.}
\label{tab:results} 
\centering
\begin{tabular}{c c c c}
\specialrule{.1em}{.05em}{.05em} 
Method & ACA & Macro-F1 & AUC\\
\hline
Bravo~\emph{et al.}~\cite{Mar2017Automatic} & 0.5051 & 0.5081 & -  \\
ResNet-50~\cite{ResNet} & 0.4820  & 0.4877 & 0.8091  \\
ResNet-50, MSE & 0.4985 & 0.4995 & 0.8144 \\
\textbf{TSBN (ResNet-50)}   & \textbf{0.5212}  & \textbf{0.5242} & \textbf{0.8218}  \\
\hline
BiRA-Net~\cite{zhao2019biranet} & 0.5431 & 0.5723 & 0.8448 \\
\textbf{TSBN (BiRA-Net)}    &\textbf{0.5513} & \textbf{0.5792}& \textbf{0.8490} \\
\specialrule{.1em}{.05em}{.05em} 
\end{tabular}

\end{table}


\begin{table}[htbp]
\caption{Grading accuracies in each DR level.}
\label{tab:per_cls_acc}
\centering

\begin{tabular}{c c c}
\specialrule{.1em}{.05em}{.05em} 
Class & ResNet-50 & \textbf{TSBN (ResNet-50)}\\
\hline
Normal (0) &0.6250  & \textbf{0.6442}  \\

Mild (1) & 0.3814 & \textbf{0.4327}  \\

Moderate (2) &0.4327  & \textbf{0.4423}  \\

Severe (3)  &0.4103  &\textbf{0.4615}   \\

Proliferative (4) & 0.5609  &\textbf{0.6250}\\
\specialrule{.1em}{.05em}{.05em} 
\end{tabular}

\end{table}
\begin{figure}

\begin{minipage}{.48\linewidth}
\centering
\subfloat[]{\label{main:a}\includegraphics[scale=.35]{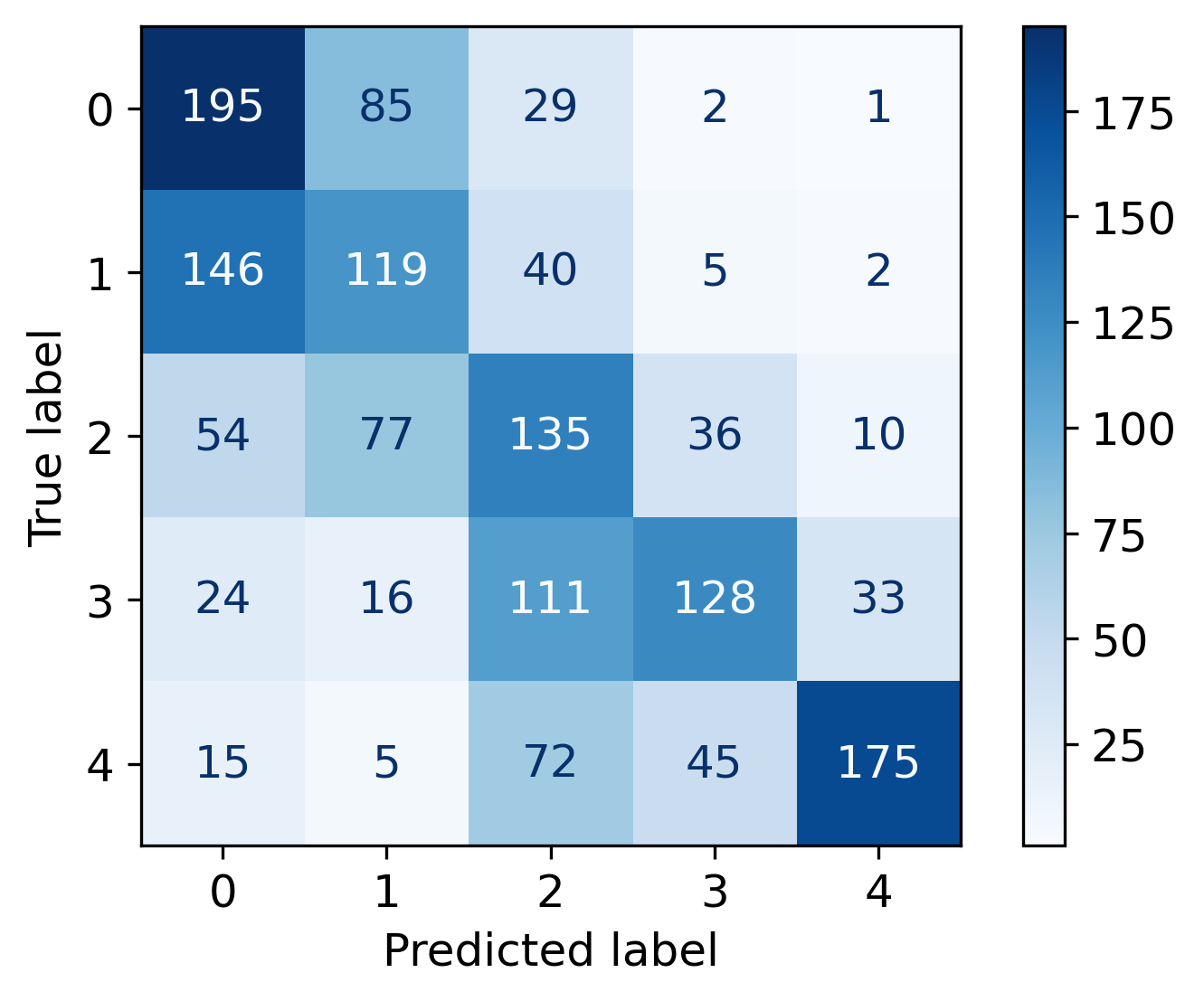}}
\end{minipage}%
\begin{minipage}{.48\linewidth}
\centering
\subfloat[]{\label{main:b}\includegraphics[scale=.35]{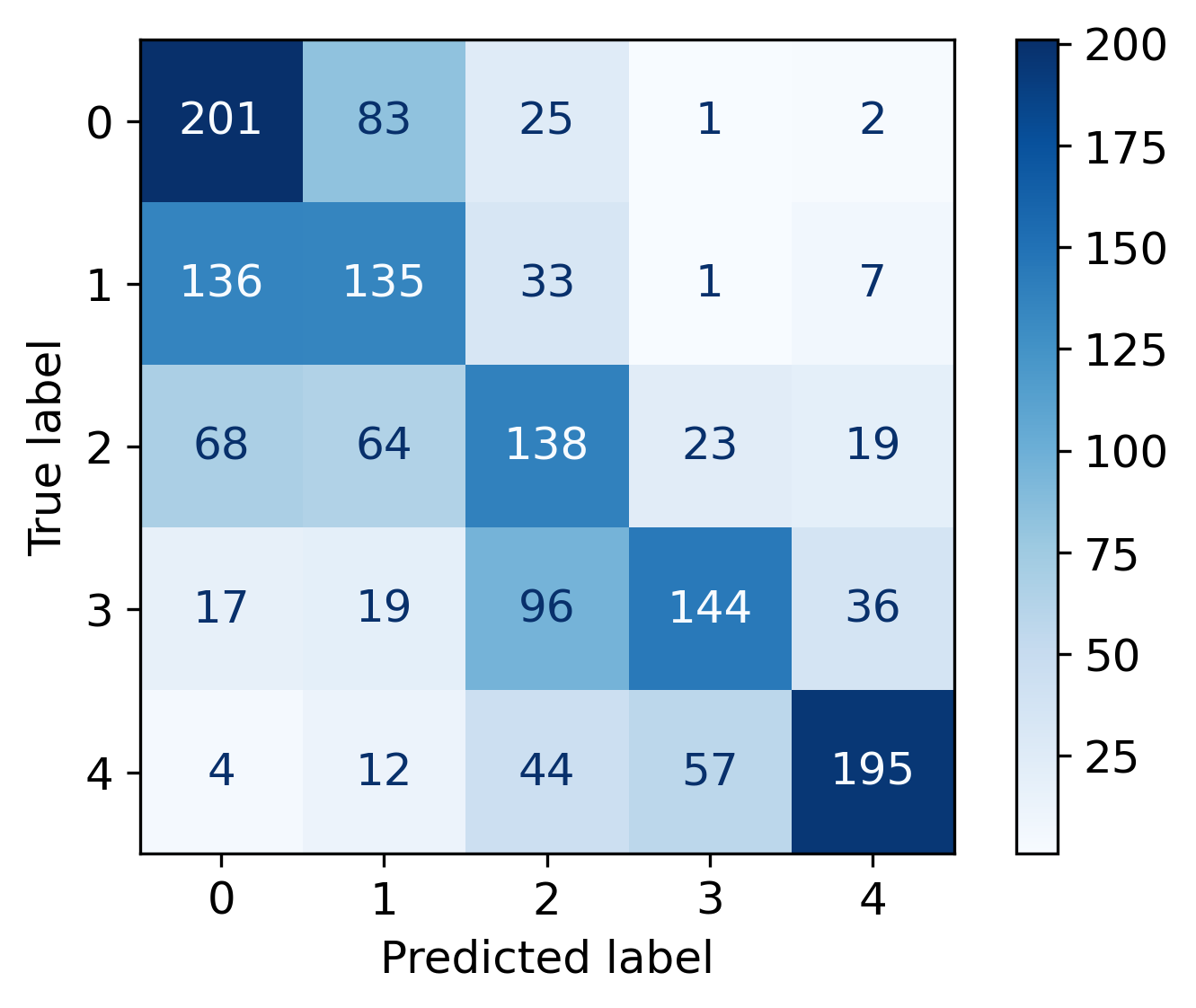}}
\end{minipage}\par\medskip
\caption{Confusion matrices of (a) ResNet-50, (b) TSBN (ResNet-50).}
\label{fig:confusion}
\end{figure}
In Table~\ref{tab:results}, the experimental results of our approach are compared with baseline methods. When using ResNet-50 as the backbone, our method has a clear advantage of $4\%$ increase in classification accuracy. It proves that our training strategy and loss function can better optimize the model. We also outperform the original BiRA-Net, a more sophisticated architecture engineered by a dedicated grading loss~\cite{zhao2019biranet}. The results confirm that our approach is widely applicable to different backbone models for DR grading. 

In Table.~\ref{tab:per_cls_acc}, it is clear that our approach reaches higher classification accuracy in all levels of DR grading, especially in higher levels where the training samples are sparse. Fig.~\ref{fig:confusion} represents confusion matrices for ResNet-50 and our method respectively. Our model distinguishes more cases among moderate to serious DR grades (grade $2$ to $4$). In other words, our model identifies patients who require clinical attention most, proving its clinical significance.

Examples of saliency maps on the test data are illustrated in Fig.~\ref{fig:gradcam} with Grad-CAM~\cite{Selvaraju_2017_ICCV}. Comparing the left and right saliency maps of the same patient, it is evident that our model is activated at symmetrical positions with similar intensities. It confirms that the similarities of DR symptoms in both eyes are informative for DR grading. It is also validated that our approach effectively learns such similarities for classification. Besides, it is observed that there are some discrepancies between the saliency maps of left and right eyes, due to variances of the DR symptoms and limitations in the model capability.   

\section{CONCLUSIONS AND FUTURE WORK}\label{section5}
We propose a two-stream binocular network, which explores similarities and correlations between left and right eyes for DR grading. The framework consists of $2$ CNNs with shared weights and classifies the left and right eyes of the same patient respectively. To capture the similarities between left and right eyes, a hybrid loss function is proposed, which combines a contrastive grading loss and a weighted cross-entropy loss. Extensive experiments has shown that our approach is effective. The left-right eye similarities are visualized in the saliency maps of our model. In the future, we can further improve the model performance by exploring left-right eye correlations with domain knowledge.

\bibliographystyle{IEEEbib}
\bibliography{refs1.bib}

\end{document}